\documentclass[journal]{IEEEtran}
\usepackage{cite}
\usepackage{amsmath}
\usepackage{comment}
\usepackage{csquotes}
\usepackage{hyperref}
\usepackage[capitalise,noabbrev]{cleveref}
\usepackage[ruled,linesnumbered]{algorithm2e}
\usepackage{graphicx}
\Crefname{figure}{\text{Fig.}}{\text{Figs.}}	

\begin{document}
%
\title{Benchmarking Autonomous Vehicles: A Driver Foundation Model Framework}
%
%
%

\author{Yuxin~Zhang,
        Cheng~Wang,
        Hubert P. H. Shum 
\thanks{Y. Zhang is with the National Key Laboratory of Automotive Chassis Integration and Bionics, Jilin University, Changchun 130025, China and the Department of Computer Science, Durham University, Durham, DH1 3LE, UK, and DRIVEResearch, Changchun, China, e-mail: yuxinzhang@jlu.edu.cn.}
\thanks{C. Wang is with the School of Engineering and Physical Sciences, Heriot-Watt University, Edinburgh, EH14 4AS, UK, e-mail:cheng.wang@hw.ac.uk.}
\thanks{H. P. H. Shum is with the Department of Computer Science, Durham University, Durham, DH1 3LE, UK, e-mail:hubert.shum@durham.ac.uk.}
}

\markboth{}%
{Shell \MakeLowercase{\textit{et al.}}: Bare Demo of IEEEtran.cls for IEEE Journals}

\maketitle

\begin{abstract}
Autonomous vehicles (AVs) are poised to revolutionize global transportation systems. However, its widespread acceptance and market penetration remain significantly below expectations. This gap is primarily driven by persistent challenges in safety, comfort, commuting efficiency and energy economy when compared to the performance of experienced human drivers. We hypothesize that these challenges can be addressed through the development of a driver foundation model (DFM). Accordingly, we propose a framework for establishing DFMs to comprehensively benchmark AVs. Specifically, we describe a large-scale dataset collection strategy for training a DFM, discuss the core functionalities such a model should possess, and explore potential technical solutions to realize these functionalities. We further present the utility of the DFM across the operational spectrum, from defining human-centric safety envelopes to establishing benchmarks for energy economy. Overall, We aim to formalize the DFM concept and introduce a new paradigm for the systematic specification, verification and validation of AVs.
\end{abstract}


%
\IEEEpeerreviewmaketitle

\section{Introduction}

\IEEEPARstart{A}{utonomous} vehicles (AVs) are expected to provide safe, clean and efficient mobility for our society. Driven by this vision, significant research and development efforts have been dedicated to advancing autonomous driving functionalities. Advanced driver assistance systems are already commercially available, the current focus has shifted toward automated driving systems to enable greater flexibility and autonomy for customers. Despite these advancements, the social acceptance of AVs remains limited. Critical safety concerns are still unaddressed, as significant milestones required to validate safety goals have yet to be reached. Beyond safety, other ride performance such as commuting efficiency, energy economy and comfort, is less focused despite its significant impact on users' driving experience at each moment while using the AV. Consequently, there is an urgent need to define a reference model for calibrating AV behavior in terms of  ride performance, to better align with human expectations.


\begin{figure}[!ht]
    \centering
    \includegraphics[width=\columnwidth]{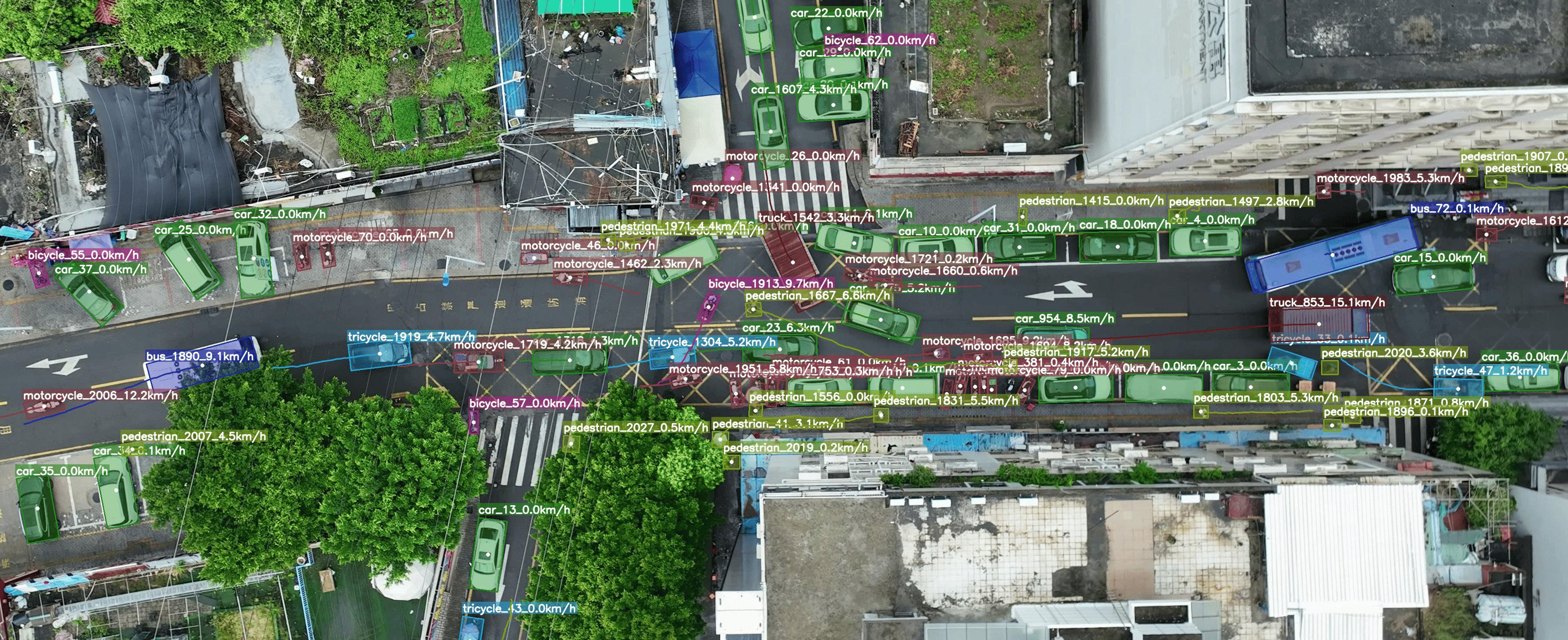}
    \caption{One example to illustrate the extracted information (e.g., object class, velocity, position) from drone-based data collection strategy.}
    \label{fig:processedsample}
\end{figure}

The most relevant work is the careful and competent driver model (CCDM)  \cite{unece_uniform_nodate} and the Responsibility-Sensitive Safety model \cite{shalev2017formal}. However, they primarily focuses on constructing a safety benchmark for AVs while leaving ride performance under considered. Additionally, such benchmark models are rule-based and support only $1 \sim 3$ agent interaction scenarios. The simplicity and these inherent limitations constraint their applicability to end-to-end AVs. Several studies \cite{olleja2025validation, wang2024safety} have demonstrated their inaccuracy and limited suitability across different  operational design domains (ODDs). As a result, current benchmark models undermine the effectiveness of safety assessment and hinders the social acceptance of AVs. 

We aim to address these limitations by creating a large scale human driver behavior dataset and conceptualizing a driver foundation model (DFM) to provide accurate and reasonable benchmarks for AVs under diverse multi-agent interactions. More specifically, this paper is grounded in large scale real world human driving data collected within a target ODD with the objective of extracting a realistic baseline of human driving capability. Rather than focusing solely on safety which can paradoxically lead to unsafe behavior at larger spatial or temporal scales when considered in isolation, the proposed approach jointly accounts for comfort, commuting efficiency and energy economy. In doing so, it aims to enhance AV safety while simultaneously improving user acceptance. This framework encompasses the construction of a database that comprehensively reflects the characteristics of the target market ODD, together with a corresponding DFM technical framework built upon this database. Finally, we demonstrate the potential of the proposed DFM as a reference benchmark for developing and evaluating AVs across multiple dimensions of performance. This evaluation includes safety and ride performance, ensuring that AVs are both human like and socially acceptable.

\begin{figure*}[!ht]
    \centering
    \includegraphics[width=\textwidth]{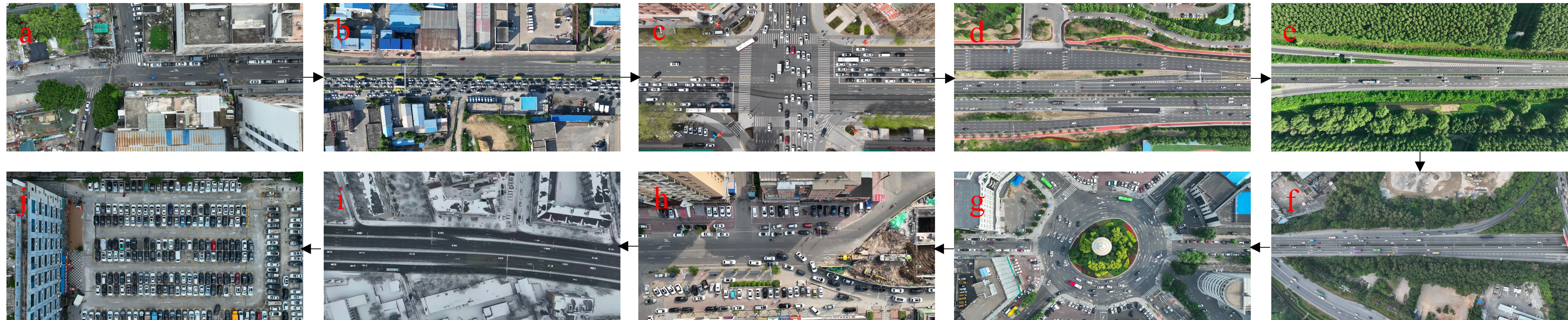}
    \caption{Our exemplary scenario coverage for an urban ODD. (a) residential apartment area; (b) urban arterial road; (c) intersections; (d) on-ramp; (e) expressway; (f) off-ramp; (g) roundabouts; (h) accident/construction zone; (i) icy and snowy road; (j) parking lot.}
    \label{fig:scenarios}
\end{figure*}

\section{Concept}
In this section, we elaborate the concept to create our DFM framework including \textit{data foundation} and \textit{framework discussion}.

\subsection{Data foundation}
Our data is collected by drone-based technique. Specifically, a drone hovers over a road section for certain time and records videos. The data processing chain is: \textit{data collection} $\rightarrow$ \textit{video correction} $\rightarrow$ \textit{object detection} $\rightarrow$ \textit{object tracking} $\rightarrow$ \textit{data smoothing}. Finally, trajectories (e.g., position, velocity and yaw) of all road users appeared in a video are obtained, as shown in \cref{fig:processedsample}. More detail descriptions of each step can be found in \cite{wang2025comprehensive}. Until January 2026, the number of our trajectories are over 7.5 millions \footnote{https://www.driveresearch.tech/}, occupying the largest aerial survey dataset in the world to the authors' best knowledge. Importantly, our dataset spans a wide range of scenarios across diverse ODDs. Taking an urban ODD as an illustrative example (see \cref{fig:scenarios}), a representative journey begins in a residential apartment area (a), where the vehicle exits local streets and merges onto an urban arterial road (b). It subsequently encounters intersections (c) before entering an expressway via an on-ramp (d), followed by expressway cruising (e) and a later off-ramp transition (f). The route further includes complex traffic structures such as roundabouts (g). Throughout this end-to-end journey, the vehicle may experience unexpected events, including traffic accidents (h) and adverse weather conditions (i), before finally reaching its destination and completing a parking maneuver (j). The intensive scenario coverage facilitates benchmarking an AV in its target ODD. Part of our data is open-sourced at \cite{zhang2023ad4che}.

Compared to autonomous driving datasets such as Waymo \cite{ettinger2021large}, our dataset provides a more suitable foundation for training a DFM because it captures unobstructed, global traffic dynamics and genuine human driving behavior at scale. The top-down aerial perspective eliminates occlusions and sensor bias inherent to ego-vehicle datasets, enabling precise, continuous observation of multi-agent interactions and long-horizon decision making. Additionally, the data preserves the natural diversity, efficiency–safety trade-offs and comfort-related maneuvers exhibited by human drivers within the target ODD. Using generative models \cite{chang2026design} to synthesize rare scenarios is also a promising direction for further enhancing data diversity.

\begin{figure*}[t]
  \centering
  \includegraphics[width=\textwidth]{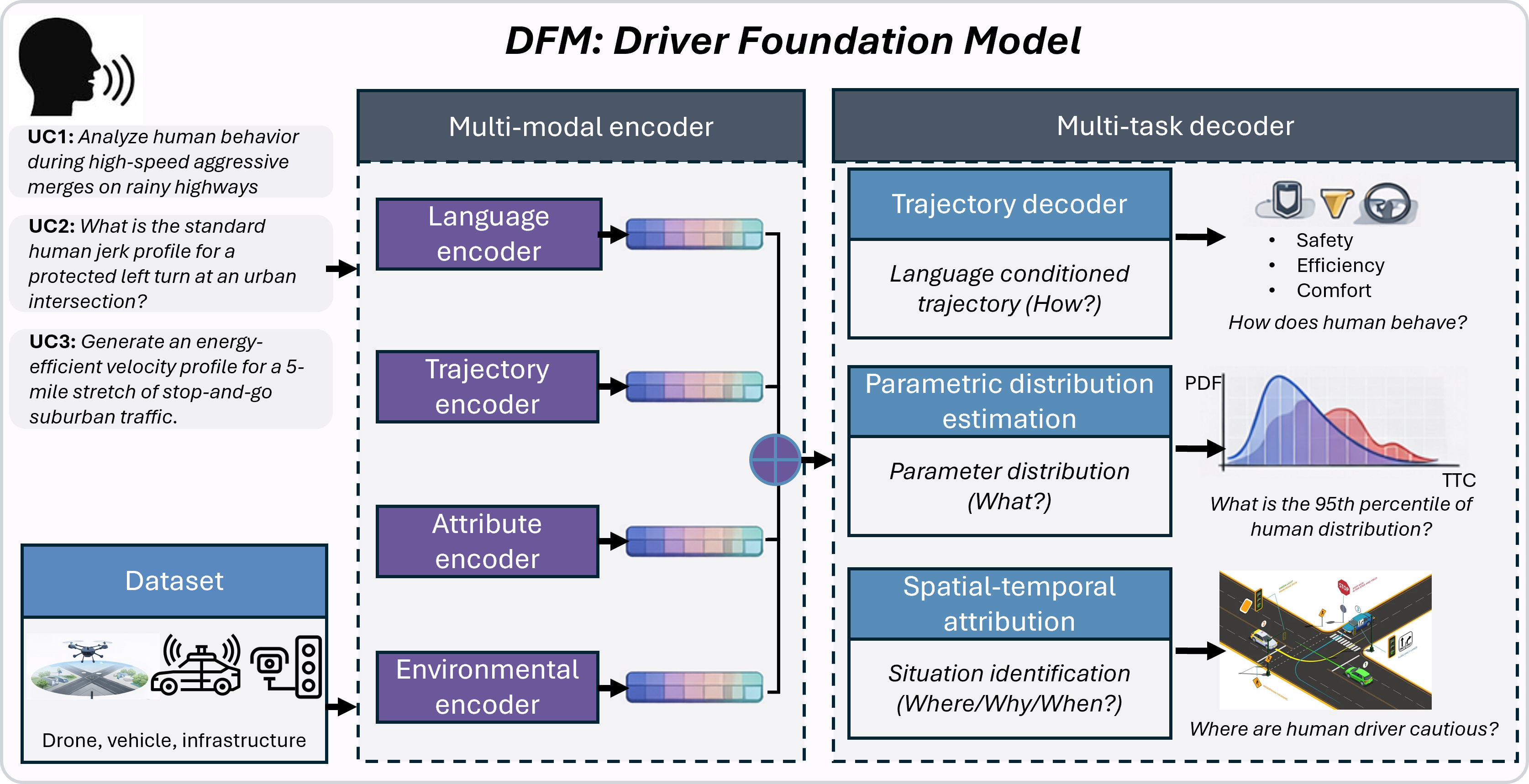}
  \caption{The proposed DFM framework. A multi-modal encoder and a multi-task decoder are the two main components to address various questions (see three typical use cases (UCs)) from users and to realize multi-functionalities for benchmarking AVs.}
  \label{fig:architecture}
\end{figure*}

\subsection{Framework discussion}
Our approach to establishing the DFM framework begins with a discussion of its desired functionalities, which serves as the basis for deriving its architectural requirements. With these requirements established, we then discuss the potential architecture for fulfilling these requirements. This ensures that the resulting model is not only technically sound but also purpose-built for the complex task of AV benchmarking. Our concept of the DFM framework is illustrated in \cref{fig:architecture}. 

\textbf{Functionalities.} We hypothesize that the DFM serves as a comprehensive oracle for AV development by addressing the \textit{"How"}, \textit{"What"}, \textit{"Where"}, \textit{"When"} and \textit{"Why"} of human behavior. This quintet of outputs transforms raw data into an actionable knowledge base. 
The "How" characterizes the physical manifestation of human behavior in a given scenario. This provides reference trajectories for AVs, facilitating safety assessment of AVs and ensuring their social acceptance to surrounding road users. 
The "What" focuses on the statistical distributions underlying human performance. it outputs the probabilistic ranges for critical parameters such as time-to-collision (TTC). This data defines the "competence envelope", allowing developers to specify performance requirements for AVs. 
The "Where" identifies the specific situational hotspots that trigger distinct human behaviors. By highlighting environmental features, such as a complex merging zone, the DFM pinpoints the geographic and topological areas of interest. This enables developers to focus their testing and sensor-tuning efforts on the zones that most significantly influence human decision-making.
The "When" addresses the timing and triggering of behavioral transitions. It identifies the critical moment or environmental conditions (e.g., bad weather) that induce a specific maneuver of a driver. For instance, under what snow conditions drivers will drive slowly. For AV developers, this provides the temporal benchmarks necessary to define trigger events and corresponding mitigation actions.
Finally, the "Why" provides an interpretative layer that identifies which specific environmental cues are most influential in a driver’s behavior. This can pinpoint the exact pieces of information, such as a distant pedestrian's posture or the specific curvature of a lane, which necessitates a maneuver. This insight is invaluable for AV developers as it assists in defining perception requirements and optimal sensor configurations. By understanding what information a human prioritizes, engineers can ensure that the AV’s sensing suite is tuned to capture the most safety-critical data in given scenes.

\textbf{Framework.} To realize the functionalities above, we hypothesize a framework that includes a \textit{multi-modal encoder} and a \textit{multi-task decoder}. The multi-modal encoder for a DFM consists of several sub-encoders. A language encoder will encode human queries into latent embeddings via a pertained large language model (LLM) backbone such as GPT \cite{sanderson2023gpt}. This provides the model with an understanding of human language, allowing it to know that terms like "aggressive" and "sudden" are semantically related when describing braking. A trajectory encoder processes kinematic data of multiple agents to produce structured motion embeddings. A attribute encoder focuses on the physicality of the agent (such as size and agent class). It ensures the model knows that a "cautious stop" for a semi-truck requires more distance than for a sedan. Finally, an environmental encoder captures external constraints like weather and lighting. This ensures the model can be adjusted according to external environmental conditions (e.g., lower speed expectations on a snowy night). To facilitate interaction between modalities, the encoded latent representations are synthesized via cross-attention bridges or joint shared latent spaces. A cross-attention approach enables dynamic alignment, where linguistic queries selectively attend to specific spatial or environmental features, while a joint shared latent space projects all modalities into a unified embedding. 

To recover the encoded information and realize the proposed functionalities, a multi-task decoder utilizing parallel feature projection is essential. In this architecture, the fused latent representation is branched into specialized, independent neural heads, each dedicated to answering a specific "W" question:
A trajectory decoder reconstructs language-conditioned, metric-space trajectories from the latent embeddings. This head serves as the generative core for benchmarking AV decisions, providing a "human-optimal" reference trajectory that accounts for the specific vehicle attributes and environmental constraints provided by the encoders.
A parametric distribution head performs key parameter distribution analysis like velocity, acceleration and jerk. The model captures not only the "typical" value for target velocity or peak deceleration but also the uncertainty or diversity of human responses in that specific context. This technique provides the mathematical boundaries of "competent" human behavior, enabling the AV developer to see if their AV's performance falls within the $5th$ or $95th$ percentile of human distribution for a specific scenario. 
A spatial-temporal attribution module performs a back-projection of attention weights onto the environment. It answers "Where" by highlighting situational hotspots and "Why" by identifying the specific causal factors, such as a slippery road segment or a merging neighbor, which necessitates a behavioral shift.

\textbf{Discussion.}
The proposed DFM framework establishes a new paradigm for AV development by transitioning from traditional, narrow imitation learning to a comprehensive, human-centric benchmarking framework. By decomposing driving behavior into the "Five Ws"—How, What, Where, When, and Why—the model leverages a multi-modal quad-encoder architecture to synthesize benchmarks that are semantically grounded and physically realistic. This framework utilizes flexible integration strategies, such as cross-attention or shared latent spaces, and a parallel multi-task decoder to provide actionable engineering outputs, ranging from kinematic reference paths to explainable diagnostic saliency. Ultimately, by augmenting large-scale empirical data with vehicle-specific and atmospheric context, the DFM closes the gap between qualitative human intent and quantitative system verification, offering a scalable solution to the challenges of safety, comfort, commuting efficiency and energy economy in AVs.

\section{DFM for ride performance}
Developing such a DFM is beneficial as it can significantly guide the design of intuitive vehicle behaviors, aid the verification and validation of safety-critical systems, and facilitate the standardized benchmarking of AVs against human performance limits. In the following, We will discuss its utility across four aspects: crossing from safety to economy. 

\textbf{Safety benchmark.} The DFM can be used for scenario-based testing by generating adjustable driving behavior for the agents surrounding an AV. By deploying different driving characteristics for the surrounding agents, the AV is facing various types of driver, such as aggressive and cautious, which can test the robustness of the AV in handling different drivers. It can address the limitations of existing CCDMs by leveraging a multi-modal transformer architecture to synthesize high-fidelity, socially-aware behaviors across diverse driving contexts. By extracting typical human driver behaviors and their associated parameter distributions, the DFM enables the derivation of formal specifications that ground AV performance in empirical human baselines. Additionally, by using the spatial attribution module, developers can pinpoint exactly where their system deviates from human performance, allowing for targeted upgrades. Consequently, the DFM facilitates both the rigorous specification of AV requirements and the accelerated verification of their safety and performance.

\textbf{Comfort benchmark.} Except safety, comfort is another important factor that affect user's acceptance of AVs. The DFM serves as a vital bridge for passenger acceptance by defining the human comfort envelope. Existing AV comfort evaluation is largely subjective, relying on developers’ perceptions to assess whether the AV behaves in a human-like manner or exhibits robotic and jerky motions. By utilizing the parametric distribution head, the DFM extracts longitudinal jerk and lateral acceleration from human drivers across specific scenarios, such as high-speed merges or urban left turns. This allows developers to design AVs that are not only safe but also naturalistic. By calibrating the AV’s motion profiles to stay within the statistical bounds of human behavior, the DFM ensures that the AV's decision-making feels intuitive to its occupants. Consequently, comfort is transformed from a subjective feeling into a quantifiable, language-gated specification that can be verified during road testing.

\textbf{Commuting efficiency benchmark.} Unlike safety-only models that may lead to overly hesitant behavior, the DFM uses the multi-task decoder to identify the "How" and "What" of human-optimal progress. By querying the model for "efficient goal-reaching in dense traffic", developers can extract the ideal velocity profiles and path-planning strategies that human drivers use to maintain a high rate of progress without compromising safety. For instance, the model can output a distribution of "standard human travel times" for a specific intersection or highway stretch under varying traffic densities. This allows AV developers to validate their AV against a human baseline, ensuring the AV does not become a "mobile bottleneck" but instead moves with the purposeful, goal-oriented flow of expert human traffic.

\textbf{Energy economy benchmark.} By querying the model for "optimal momentum conservation" in complex ODDs, the DFM provides reference velocity profiles that minimize unnecessary braking and acceleration cycles.
The model identifies how expert human drivers utilize the road's geometry and traffic flow to maintain kinetic energy, providing a benchmark for "Eco-Driving". This is particularly important for heavy-duty trucks, where potential energy consumption is substantial. The DFM can output distributions of acceleration and deceleration rates that lead to optimal energy efficiency for autonomous trucks. By comparing the AV's current energy footprint against human efficiency baselines, developers can tune powertrain systems to maximize vehicle range and reduce operational costs, ensuring AVs operate at the intersection of high traffic throughput and minimal resource consumption.

\section{Conclusion}
This position paper contributes to future mobility by proposing the DFM framework for comprehensive benchmarking of AVs using large-scale datasets, with the goal of enhancing both social and technical acceptance of AVs. In the era of AI,  our drone-based data collection strategy provides a significant complement to existing data acquisition approaches, with the potential to further improve the performance of AVs. We argue that drone-based data collection offers substantial value and untapped potential. In addition, we observe that a DFM serving as a reference model for AV evaluation is currently missing, and that the need for such a model becomes increasingly urgent when considering the broader social impacts of AV deployment. Given this importance, we aim to further concretize the technical roadmap for DFMs and to work jointly with stakeholders to realize this vision.

\ifCLASSOPTIONcaptionsoff
  \newpage
\fi



\bibliographystyle{IEEEtran}
\bibliography{bibtex/reference}
%




\end{document}